\pdfoutput=1

\documentclass[11pt]{article}

\usepackage[]{acl}

\usepackage{times}
\usepackage{latexsym}

\usepackage[T1]{fontenc}

\usepackage[utf8]{inputenc}
\usepackage{graphicx}
\usepackage{microtype}
\usepackage{tikz}
\usepackage{amsmath}
\usepackage{subcaption}

\usepackage{mathpartir}
\usetikzlibrary{positioning}
\usetikzlibrary{calc}
\usetikzlibrary{decorations.pathreplacing}
\usepackage{forest}
\forestset{
dg edges/.style={for tree={parent anchor=south, child anchor=north,align=center,base=bottom,where n children=0{tier=word,edge=dotted,calign with current edge}{}}},
}

\usepackage{cleveref}
\crefformat{subsection}{\S#2#1#3}

\usepgflibrary{arrows.meta}
\usetikzlibrary{fit}
\usepackage{multirow, multicol}
\usepackage{pgfplots}

\usepackage{enumitem}
\usepackage{booktabs}       

\setlist[itemize]{noitemsep}

\mathchardef\mhyphen="2D

\DeclareDocumentCommand{\trapezoid}{O{2.0} O{1.0} O{0.5} m m}{
    \begin{scope}[scale=0.9,thick]
        \draw[anchor=mid] (0, 0) -- (0, -{#2}) node[below=9pt,anchor=base] {\ensuremath{#4}} -- ({#1}, -{#2}) node [below=9pt,anchor=base] {\ensuremath{#5}} -- ({#1}, -{#3}) -- cycle;
    \end{scope}
}

\DeclareDocumentCommand{\trapezoidd}{O{2.0} O{1.0} O{0.5} m m m}{
    \begin{scope}[scale=0.9,thick]
        \draw[anchor=mid] (0, 0) -- (0, -{#2}) node[below=9pt,anchor=base] {\ensuremath{#4}} -- ({#1}, -{#2}) node [below=9pt,anchor=base] {\ensuremath{#5}} -- ({#1}, -{#3}) -- cycle;
            \draw node[] at ( $({#2}, 0.2)!0.5!(0, 0.2)$) {\ensuremath{#6}};
    \end{scope}
}

\DeclareDocumentCommand{\trapezoiddd}{O{2.0} O{1.0} O{0.5} m m m m}{
    \begin{scope}[scale=0.9,thick]
        \draw[anchor=mid] (0, 0) -- (0, -{#2}) node[below=9pt,anchor=base] {\ensuremath{#4}} -- ({#1}, -{#2}) node [below=9pt,anchor=base] {\ensuremath{#5}} -- ({#1}, -{#3}) -- cycle;
            \draw node[] at ( $({#2}, 0.7)!0.5!(0, 0.7)$) {\ensuremath{#6}};
            \draw node[] at ( $({#2}, 0.2)!0.5!(0, 0.2)$) {\ensuremath{#7}};
    \end{scope}
}

\DeclareDocumentCommand{\trapezoiddleft}{O{2.0} O{1.0} O{0.5} m m m}{
    \begin{scope}[scale=0.9,thick]
        \draw[anchor=mid] (0, 0) -- (0, -{#2}) node[below=9pt,anchor=base] {\ensuremath{#4}} -- ({#1}, -{#2}) node [below=9pt,anchor=base] {\ensuremath{#5}} -- ({#1}, -{#3}) -- cycle;
            \node at ( $({#2}, 0.5)!0.5!(0, 0.5)$) [] {\ensuremath{#6}};
    \end{scope}
}

\DeclareDocumentCommand{\trapezoiddleftt}{O{2.0} O{1.0} O{0.5} m m m m}{
    \begin{scope}[scale=0.9,thick]
        \draw[anchor=mid] (0, 0) -- (0, -{#2}) node[below=9pt,anchor=base] {\ensuremath{#4}} -- ({#1}, -{#2}) node [below=9pt,anchor=base] {\ensuremath{#5}} -- ({#1}, -{#3}) -- cycle;
            \node at ( $({#2}, 1)!0.5!(0, 1)$) [] {\ensuremath{#6}};
                        \node at ( $({#2}, 0.5)!0.5!(0, 0.5)$) [] {\ensuremath{#7}};
    \end{scope}
}

\DeclareDocumentCommand{\righttriangle}{O{0.5} O{0.5} m m }{
    \begin{scope}[scale=0.9,thick]
        \draw[anchor=mid] (0, 0) -- (0, -{#2}) node[below=9pt,anchor=base] {\ensuremath{#3}} -- ({#1}, -{#2}) node [below right=9pt and 3pt,anchor=base] {\ensuremath{#4}} -- cycle;
    \end{scope}
}

\DeclareDocumentCommand{\righttrianglee}{O{0.5} O{0.5} m m m}{
    \begin{scope}[scale=0.9,thick]
        \draw[anchor=mid](0, 0) -- (0, -{#2}) node[below=9pt,anchor=base] {\ensuremath{#3}} -- ({#1}, -{#2}) node [below right=9pt and 3pt,anchor=base] {\ensuremath{#4}} -- cycle;
        \draw node[] at ( $({#1}, 0.2)!0.5!(0, 0.2)$) {\ensuremath{#5}};
    \end{scope}
}

\DeclareDocumentCommand{\lefttriangle}{O{0.5} O{0.5} m m}{
    \begin{scope}[scale=0.9,thick]
        \draw[anchor=mid] (0, -{#2}) node[below left=9pt and 3pt,anchor=base] {\ensuremath{#3}} -- ({#1}, -{#2}) node [below=9pt,anchor=base] {\ensuremath{#4}} -- ({#1}, 0) -- cycle;
    \end{scope}
}

\DeclareDocumentCommand{\square}{O{0.5} O{0.5} m m m}{
    \begin{scope}[scale=0.9,thick]
        \draw[anchor=mid] (0, -{#2}) node[below left=9pt and 3pt,anchor=base] {\ensuremath{#3}} -- ({#1}, -{#2}) node [below=9pt,anchor=base] {\ensuremath{#4}} -- ({#1}, 0) --  (0, 0) -- cycle;
                \draw node[] at ( $({#1}, 0.2)!0.5!(0, 0.2)$) {\ensuremath{#5}};
    \end{scope}
}

\DeclareDocumentCommand{\leftclosetriangle}{O{0.5} O{0.5} m m}{
    \begin{scope}[scale=0.9,thick]
        \draw[anchor=mid] (0, -{#2}) node[below left=9pt and 3pt,anchor=base] {\ensuremath{#3}} -- ({#1}, -{#2}) node [below=9pt,anchor=base] {\ensuremath{#4}} -- ({#1}, 0) -- cycle;
        \draw[anchor=mid] ({#1}, -{#2}-0.05) -- (0, -{#2}-0.05);
    \end{scope}
}

\DeclareDocumentCommand{\leftclosetrianglee}{O{0.5} O{0.5} m m m}{
    \begin{scope}[scale=0.9,thick]
        \draw[anchor=mid] (0, -{#2}) node[below left=9pt and 3pt,anchor=base] {\ensuremath{#3}} -- ({#1}, -{#2}) node [below=9pt,anchor=base] {\ensuremath{#4}} -- ({#1}, 0) -- cycle;
        \draw[anchor=mid] ({#1}, -{#2}-0.05) -- (0, -{#2}-0.05);
        \draw node[] at ( $({#1}, 0.2)!0.5!(0, 0.2)$) {\ensuremath{#5}};
    \end{scope}
}

\DeclareDocumentCommand{\lefttrianglee}{O{0.5} O{0.5} m m m}{
    \begin{scope}[scale=0.9,thick]
        \draw[anchor=mid] (0, -{#2}) node[below left=9pt and 3pt,anchor=base] {\ensuremath{#3}} -- ({#1}, -{#2}) node [below=9pt,anchor=base] {\ensuremath{#4}} -- ({#1}, 0) -- cycle;
        \draw node[] at ( $({#1}, 0.2)!0.5!(0, 0.2)$) {\ensuremath{#5}};
    \end{scope}
}

\DeclareDocumentCommand{\rightclosetriangle}{O{0.5} O{0.5} m m}{
    \begin{scope}[scale=0.9,thick]
        \draw[anchor=mid] (0, 0) -- (0, -{#2}) node[below=9pt,anchor=base] {\ensuremath{#3}} -- ({#1}, -{#2}) node [below right=9pt and 3pt,anchor=base] {\ensuremath{#4}} -- cycle;
        \draw[anchor=mid] ({0}, -{#2}-0.05) -- ({#1}, -{#2}-0.05);

    \end{scope}
}

\DeclareDocumentCommand{\rightclosetrianglee}{O{0.5} O{0.5} m m m}{
    \begin{scope}[scale=0.9,thick]
        \draw[anchor=mid] (0, 0) -- (0, -{#2}) node[below=9pt,anchor=base] {\ensuremath{#3}} -- ({#1}, -{#2}) node [below right=9pt and 3pt,anchor=base] {\ensuremath{#4}} -- cycle;
        \draw[anchor=mid] ({0}, -{#2}-0.05) -- ({#1}, -{#2}-0.05);
        \draw node[] at ( $({#1}, 0.2)!0.5!(0, 0.2)$) {\ensuremath{#5}};
    \end{scope}
}

\DeclareDocumentCommand{\triangle}{O{0.5} O{0.5} m m m m}{
    \begin{scope}[scale=0.9,thick]
        \draw[anchor=mid] (0, 0) node[draw, circle, black, fill, scale=0.6]{} -- (-{#1}, -{#2}) node[below=9pt,anchor=base] {\ensuremath{#3}}  --  (0, -{#2}) node[below=9pt,anchor=base] {\ensuremath{#4}} -- ({#1}, -{#2}) node [below right=9pt and 3pt,anchor=base] {\ensuremath{#5}} -- cycle;
        \draw node at ( 0, 0.5) {\ensuremath{#6}};
    \end{scope}
}

\DeclareDocumentCommand{\triangleclose}{O{0.5} O{0.5} m m m m}{
    \begin{scope}[scale=0.9,thick]
        \draw[anchor=mid] (0, 0) node[draw, circle, black, fill, scale=0.6]{} -- (-{#1}, -{#2}) node[below=9pt,anchor=base] {\ensuremath{#3}}  --  (0, -{#2}) node[below=9pt,anchor=base] {\ensuremath{#4}} -- ({#1}, -{#2}) node [below right=9pt and 3pt,anchor=base] {\ensuremath{#5}} -- cycle;
        \draw[anchor=mid] (-{#1}, -{#2}-0.05) -- ({#1}, -{#2}-0.05);
        \draw node[] at ( 0, 0.5) {\ensuremath{#6}};
    \end{scope}
}

\DeclareDocumentCommand{\triangleleft}{O{0.5} O{0.5} O{0.5} m m m m}{
    \begin{scope}[scale=0.9,thick]
        \draw[anchor=mid] (0, 0) node[draw, circle, black, fill, scale=0.6](head){} -- (-{#1}, -{#2}) node[below=9pt,anchor=base] {\ensuremath{#4}}  --  (0, -{#2}) node[below=9pt,anchor=base] {\ensuremath{#5}} -- ({#1}, -{#2}) node [below right=9pt and 3pt,anchor=base] {\ensuremath{#6}} -- cycle;
        \draw node[draw, circle, black, fill, scale=0.4] at (-{#3}, -{#2}) (node1) {};
        \draw node[below=9pt, anchor=base] at (-{#3}, -{#2}) {$h$};
                \draw[anchor=mid] (-{#1}, -{#2}-0.05) -- ({#1}, -{#2}-0.05);
        \draw[->] (node1) to[out=90] (head);
        \draw node[] at ( 0, 0.5) {\ensuremath{#7}};
        
    \end{scope}
}

\DeclareDocumentCommand{\triangleright}{O{0.5} O{0.5} O{0.5} m m m m}{
    \begin{scope}[scale=0.9,thick]
        \draw[anchor=mid] (0, 0) node[draw, circle, black, fill, scale=0.6](head){} -- (-{#1}, -{#2}) node[below=9pt,anchor=base] {\ensuremath{#4}}  --  (0, -{#2}) node[below=9pt,anchor=base] {\ensuremath{#5}} -- ({#1}, -{#2}) node [below right=9pt and 3pt,anchor=base] {\ensuremath{#6}} -- cycle;
        \draw node[draw, circle, black, fill, scale=0.4] at ({#3}, -{#2}) (node1) {};
        \draw node[below=9pt, anchor=base] at ({#3}, -{#2}) {$h$};
                \draw[anchor=mid] (-{#1}, -{#2}-0.05) -- ({#1}, -{#2}-0.05);
        \draw[->] (node1) to[in=45,out=90] (0,0);
        \draw node[] at ( 0, 0.5) {\ensuremath{#7}};
        
    \end{scope}
}

\title{Combining (Second-Order) Graph-Based and Headed-Span-Based Projective Dependency Parsing}

\author{Songlin Yang, Kewei Tu\thanks{\; Corresponding Author}\\
  School of Information Science and Technology, ShanghaiTech University \\
    Shanghai Engineering Research Center of Intelligent Vision and Imaging\\ 
    {\tt \{yangsl,tukw\}@shanghaitech.edu.cn}\\
 }

\begin{document}
\maketitle
\begin{abstract}
Graph-based methods, which decompose the score of a dependency tree into scores of dependency arcs, are popular in dependency parsing for decades. Recently, \citet{Yang2022Span} propose a headed-span-based method that decomposes the score of a dependency tree into scores of headed spans. They show improvement over first-order graph-based methods. However, their method does not score dependency arcs at all, and dependency arcs are implicitly induced by their cubic-time algorithm, which is possibly sub-optimal since modeling dependency arcs is intuitively useful. In this work, we aim to combine graph-based and headed-span-based methods, incorporating both arc scores and headed span scores into our model. First, we show a direct way to combine with $O(n^4)$ parsing complexity. To decrease complexity, inspired by the classical head-splitting trick, we show two $O(n^3)$ dynamic programming algorithms to combine first- and second-order graph-based and headed-span-based methods. Our experiments on PTB, CTB, and UD show that combining first-order graph-based and headed-span-based methods is effective. We also confirm the effectiveness of second-order graph-based parsing in the deep learning age,
however, we observe marginal or no improvement when combining second-order graph-based and headed-span-based methods \footnote{Our code is publicly available at \url{https://github.com/sustcsonglin/span-based-dependency-parsing}}. 



\end{abstract}

\section{Introduction}
\begin{figure}[tb!]
\centering
\scalebox{0.7}{
\begin{forest}
dg edges
[reads
  [child
    [the [the, name=x1] ]
     [child, name=x2] ]
  [reads, name=x3]
  [book
    [a [a, name=x4] ]
    [book, name=x5] ] ]
\foreach \i in {2, ..., 5}{
    \pgfmathtruncatemacro{\j}{\i-1}
\node (boundary\j) [below =1ex of x1] at  ( $ (x\j.east)!0.5!(x\i.west) $ ) {};
}
\node (boundary0) [left=4ex of boundary1] {}[];
\node (boundary5) [right=4ex of boundary4] {}[];
\draw (boundary0) --  (boundary1) node [midway, below, blue] {the};
\draw (boundary3) -- (boundary4) node [midway, below, blue] {a};
\draw ($(boundary0)+(0,-4ex)$) -- ($(boundary2)+(0,-4ex)$) node [midway, below, blue] {child};
\draw ($(boundary3)+(0,-4ex)$) -- ($(boundary5)+(0,-4ex)$) node [midway, below, blue] {book};
\draw ($(boundary0)+(0,-8ex)$) -- ($(boundary5)+(0,-8ex)$) node [midway, below, blue] {reads};
\end{forest}
}
\caption{An example projective dependency parse tree with all its headed spans.}
\label{figure:example}
\end{figure}
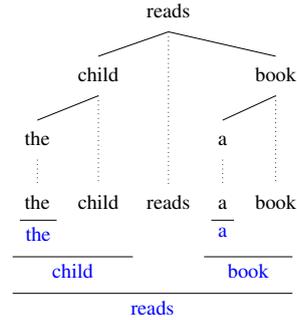

Dependency parsing is an important task in natural language processing.
There are many methods to tackle projective dependency parsing.
  In this paper, we focus on two kinds of \textit{global methods}: graph-based and headed-span-based methods. They both score all parse trees and globally find the highest scoring tree. The difference between the two is how they score dependency trees.  The simplest first-order graph-based methods \cite{mcdonald-etal-2005-online} decompose the score of a dependency tree into the scores of dependency arcs. Second-order graph-based methods \cite{mcdonald-pereira-2006-online} additionally score adjacent siblings, i.e., pairs of adjacent arcs with a shared head. There are many other higher-order graph-based methods \cite{carreras-2007-experiments, koo-collins-2010-efficient, ma-zhao-2012-fourth}.  In contrast, the headed-span-based method \cite{Yang2022Span} decomposes the score of a dependency tree into the scores of \textit{headed spans}: in a projective tree, a headed span is a word-span pair such that the subtree rooted at the word covers the span in the surface order. 
  Fig. \ref{figure:example} shows an example projective parse tree and all its headed spans.
  
First-order graph-based parsers have difficulties in incorporating sufficient subtree information before the deep learning era. \citet{DBLP:conf/iclr/DozatM17} show that first-order graph-based parsers with neural encoders and biaffine scorers can obtain high parsing accuracy.
\citet{falenska-kuhn-2019-non} argue that powerful neural encoders---such as BiLSTMs \cite{hochreiter1997long}---can encode rich subtree information implicitly, questioning the utility of high-order features. However, recent works found that high-order graph-based methods can outperform first-order graph-based methods \cite{fonseca-martins-2020-revisiting,zhang-etal-2020-efficient,wang-tu-2020-second} even with powerful neural encoders, indicating the insufficient subtree modeling of first-order graph-based methods.
 To encode more subtree information, in contrast to the line of work on higher-order parsing, \citet{Yang2022Span} choose to model headed spans, which consist of all words within the corresponding subtrees. Thus their model can utilize more subtree information than first-order graph-based methods. However, to retain the cubic parsing complexity, they abandon modeling arcs as the parsing complexity would be $O(n^4)$ otherwise (\cref{parsing1}). Modeling dependency arcs can capture the direct interactions between two words and is thus useful. Therefore, it is intuitively helpful to combine first-order graph-based and headed-span-based methods.

To decrease the parsing complexity, inspired by the classical head-splitting trick \cite{eisner-1997-bilexical}, we propose to decompose the score of a headed span into two terms, assuming that the score of the left span boundary is independent of that of the right span boundary for each headword. This allows us to adapt the Eisner algorithm to parse in cubic time considering both arc and headed span scores (\cref{parsing2}) at the cost of imposing a stronger independence assumption. More interestingly, we can also combine second-order graph-based and headed-span-based methods and need only cubic time to parse (\cref{parsing3}), which would be much slower (to the best of our knowledge, $O(n^7)$) if we do not apply the head-splitting trick.

We conduct extensive experiments on PTB, CTB, and UD. We find that combining first-order graph-based and headed-span-based methods is effective,
and applying the head-splitting trick or not result in a similar performance, thus it is more advantageous to apply this trick to enjoy a lower parsing complexity. We also confirm the effectiveness of second-order parsing in the deep learning age, however, we observe only marginal improvement or even no improvement when combining second-order graph-based and headed-span-based methods. 

\section{Scoring and Learning}
\subsection{Scoring}
\label{scoring}

 Given an input sentences $x_1, ..., x_n$, we add <bos> (beginning of sentence) and <eos> (end of sentence) as $x_0$ and $x_{n+1}$. We apply mean-pooling at the last layer of BERT \cite{devlin-etal-2019-bert} (i.e., averaging all subwords embeddings) to obtain the word-level embeddings $e_i$\footnote{For some datasets requiring the use of gold POS tags, we additionally concatenate the POS tag embedding to obtain $e_i$}. Then we feed $e_0, ..., e_{n+1}$ into a three-layer BiLSTM network to get $c_0, ..., c_{n+1}$, where $c_i = [f_i; b_i]$, $f_i$ and $b_i$ are the forward and backward hidden states of the last BiLSTM layer at position $i$ respectively. We use $h_k=[f_k, b_{k+1}]$ to represent the $k$th boundary lying between $x_k$ and $x_{k+1}$, and use $e_{i,j}=h_j-h_{i-1}$ to represent span $(i,j)$ from position $i$ to $j$ inclusive where $1\le i\le j\le n$. Then we compute:
 \begin{itemize}
     \item  $s^{\text{arc}}_{i,j}$ (for arc $x_i \rightarrow x_j$, used in all three models) by feeding  $c_i, c_j$ into a deep biaffine function \cite{DBLP:conf/iclr/DozatM17}.
     \item $s^{\text{span}}_{i,j,k}$ (for headed-span $(i, j, k)$ where $x_k$ is the headword of span $(i,j)$, used in \cref{parsing1}) by feeding $e_{i, j}, c_k$ to a deep biaffine function.
    \item $s^{\text{left}}_{k,i}$ and  $s^{\text{right}}_{k,j}$ (for headed-span $(i, j, k)$, used in \cref{parsing2} and \cref{parsing3}) by feeding $c_k, h_{i-1}$ and $c_k, h_{j}$ into two different deep biaffine functions.
    \item  $s^{\text{sib}}_{i,j,k}$ (for adjacent siblings $x_i \rightarrow \{x_j,x_k\}$ with $k<j<i$ or $i<j<k$, used in \cref{parsing3}) by feeding $c_i, c_k, c_j$ into a deep triaffine function \cite{zhang-etal-2020-efficient}. 
 \end{itemize}

\subsection{Learning}
\label{learning}
We decompose the training loss $L$ into $L_{\text{parse}} + L_{\text{label}}$. For $L_{\text{parse}}$, we use the max-margin loss  \cite{taskar-etal-2004-max}:
\begin{equation}
  L_{\text{parse}} = \max (0, \max _{y^{\prime} \neq y}(s(y^{\prime}) + \Delta(y^{\prime}, y) - s(y))  
  \label{loss}
\end{equation}
where $\Delta$ measures the difference between the incorrect tree and gold tree $y$. Here we let $\Delta$ to be the Hamming distance (i.e., the total number of mismatches of arcs, sibling pairs, and (split) headed-spans depending on the setting). We use cost-augmented inference \cite{DBLP:conf/icml/TaskarCKG05} to compute Eq. \ref{loss}, which involves the use of parsing algorithms described in the next section.  We use the same label loss $L_{\text{label}}$ in \citet{DBLP:conf/iclr/DozatM17}.

\section{Parsing}
We use the parsing-as-deduction framework \cite{pereira-warren-1983-parsing} to describe the parsing algorithms of our proposed models. 

\begin{figure*}[!ht]
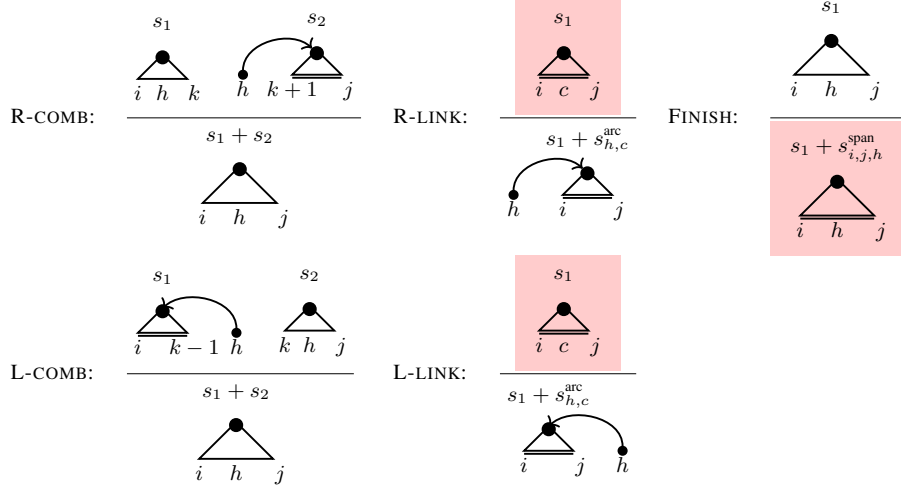

    \begin{subfigure}{\textwidth}
                    \centering
        \small 
        \scalebox{0.9}{
        \begin{tabular}{llllll}        
    \textsc{R-comb}:
    &
        $\inferrule{\tikz[]{\triangle[0.4][0.4]{i}{h}{k}{s_1}} \quad
    \tikz[]{\triangleleft[0.4][0.4][1.2]{k+1}{}{j}{s_2}}   } {
    \tikz[]{\triangle[0.6][0.6]{i}{h}{j}{s_1 + s_2}}
    }$
    &
    \textsc{R-link}:
    &
        $\inferrule{
        \colorbox{red!20}{
    \tikz[]{\triangleclose[0.4][0.4]{i}{c}{j}{s_1}} }} {
    \tikz[]{\triangleleft[0.4][0.4][1.2]{i}{}{j}{s_1 + s^{\text{arc}}_{h,c}}
    }}$
    & 
    \textsc{Finish}:
        &
        $\inferrule{
    \tikz[]{\triangle[0.6][0.6]{i}{h}{j}{s_1}} } {
    \colorbox{red!20}{
    \tikz[]{\triangleclose[0.6][0.6]{i}{h}{j}{s_1+s_{i, j, h}^{\text{span}}
    }
    }}}$
	\\    
		
	\textsc{L-comb}:
	&
	    $\inferrule{\tikz[]{\triangleright[0.4][0.4][1.2]{i}{}{k-1}{s_1}} \quad
    \tikz[]{\triangle[0.4][0.4]{k}{h}{j}{s_2}}   } {
    \tikz[]{\triangle[0.6][0.6]{i}{h}{j}{s_1 + s_2}}
    }$
    &
    \textsc{L-link}:
    &
     $\inferrule{
        \colorbox{red!20}{
    \tikz[]{\triangleclose[0.4][0.4]{i}{c}{j}{s_1}} }
    } {
    \tikz[]{\triangleright[0.4][0.4][1.2]{i}{}{j}{s_1 + s_{h,c}^{\text{arc}}}
    }}$    
	      \end{tabular}
	      }
	    \end{subfigure}
    \caption{
    Deduction rules for our modified Eisner-Satta algorithm \cite{eisner-satta-1999-efficient}. Our modifications are highlighted in red. All deduction items are annotated with their scores. Note that ``finished'' spans are marked by double underlines, whereas ``unfinished' spans take the original triangle notations.
    }
    \label{fig:algo}

\end{figure*}

\begin{figure*}[tb!]
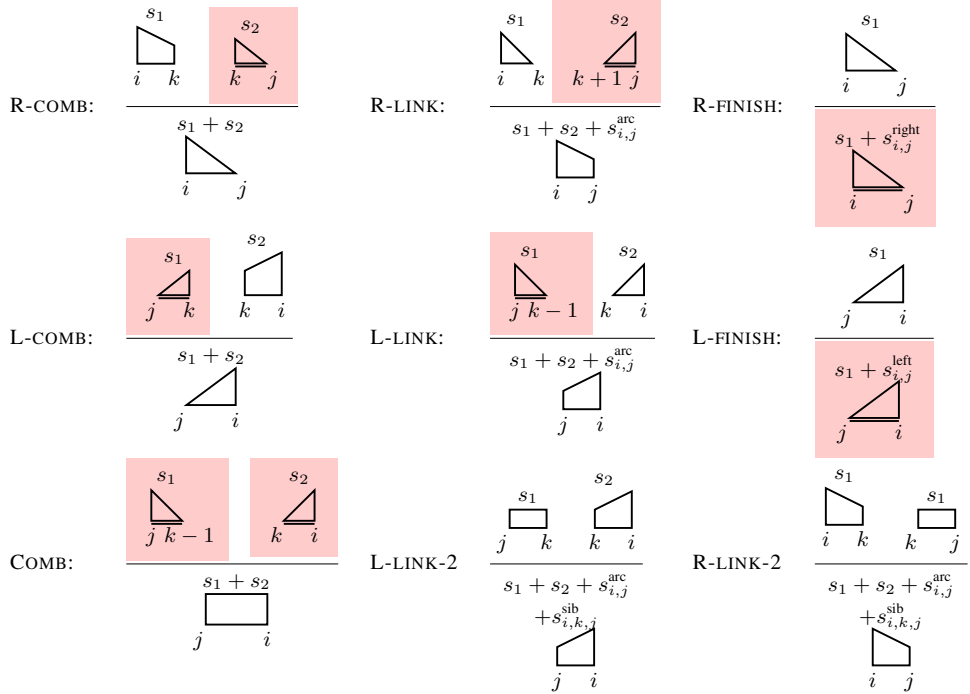

    \begin{subfigure}{\textwidth}
        \centering
        \small 
        \scalebox{0.9}{
        \begin{tabular}{llllll}        
            \textsc{R-comb}:
            &
            $\inferrule{
                \tikz[baseline=-15pt]{\trapezoidd[0.6][0.6][0.3]{i}{k}{s_1}} \quad
                \colorbox{red!20}{
                \tikz[baseline=-10pt]{\rightclosetrianglee[0.5][0.4]{k}{j}{s_2}}
                }
            } {
                \tikz[baseline=-10pt]{\righttrianglee[0.8][0.6]{i}{j}{s_1+s_2}}
            }$
            &
            \textsc{R-link}:
            &
            $\inferrule{
                \tikz[baseline=-9pt]{\righttrianglee[0.5][0.5]{i}{k}{s_1}}
                \colorbox{red!20}{
                \tikz[baseline=-9pt]{\leftclosetrianglee[0.5][0.5]{k+1}{j}{s_2}}
                }
            } {
                \tikz[baseline=-10pt]{\trapezoidd[0.6][0.6][0.3]{i}{j}{s_1+s_2+s^{\text{arc}}_{i, j}}}
            }$
            &
                        \textsc{R-finish}:
            &
            $\inferrule{
                \tikz[baseline=-10pt]{\righttrianglee[0.8][0.6]{i}{j}{s_1}} 
            } {
            \colorbox{red!20}{
                \tikz[baseline=-10pt]{\rightclosetrianglee[0.8][0.6]{i}{j}{s_1 + s^{\text{right}}_{i, j}}}}
            }$
            \\
            \textsc{L-comb}:
            &
            $\inferrule{
            \colorbox{red!20}{
                \tikz[baseline=-9pt]{\leftclosetrianglee[0.5][0.4]{j}{k}{s_1}}}\quad
                \tikz[baseline=-9pt]{\trapezoiddleft[0.6][0.4][-0.3]{k}{i}{s_2}}
            }{
                \tikz[baseline=-10pt]{\lefttrianglee[0.8][0.6]{j}{i}{s_1+s_2}}
            }$
            &
            \textsc{L-link}:
            &
            $\inferrule{
            \colorbox{red!20}{
                \tikz[baseline=-9pt]{\rightclosetrianglee[0.5][0.5]{j}{k-1}{s_1}}}
                \tikz[baseline=-9pt]{\lefttrianglee[0.5][0.5]{k}{i}{s_2}}
            } {
                \tikz[baseline=-6pt]{\trapezoiddleft[0.6][0.3][-0.3]{j}{i}{s_1+s_2+s^{\text{arc}}_{i,j}}}
            }$
            &
                                                \textsc{L-finish}:
                &
            $\inferrule{
                \tikz[baseline=-10pt]{\lefttrianglee[0.8][0.6]{j}{i}{s_1}}
            } 
            {
            \colorbox{red!20}{
                \tikz[baseline=-10pt]{\leftclosetrianglee[0.8][0.6]{j}{i}{s_1 + s^{\text{left}}_{i, j}}}
            }}$
            \\
		          \textsc{Comb}:
        &
                        $\inferrule{
            \colorbox{red!20}{
                \tikz[baseline=-9pt]{\rightclosetrianglee[0.5][0.5]{j}{k-1}{s_1}}} \quad
                \colorbox{red!20}{
                \tikz[baseline=-9pt]{\leftclosetrianglee[0.5][0.5]{k}{i}{s_2}}
                }
            } {
                \tikz[baseline=-6pt]{\square[1][0.5]{j}{i}{s_1+s_2}}
            }$
            &
        \textsc{L-link-2}
        &
        $\inferrule{
                \tikz[baseline=-6pt]{\square[0.6][0.3]{j}{k}{s_1}} \quad
                \tikz[baseline=-6pt]{\trapezoiddleft[0.6][0.3][-0.3]{k}{i}{s_2}}
            } {
                \tikz[baseline=-6pt]{\trapezoiddleftt[0.6][0.3][-0.3]{j}{i}{
                    s_1+s_2+s^{\text{arc}}_{i,j}}{+ s^{\text{sib}}_{i, k, j}}}}$
        &
        \textsc{R-link-2}
        &
        $\inferrule{
   \tikz[baseline=-15pt]{\trapezoidd[0.6][0.6][0.3]{i}{k}{s_1}} \quad \tikz[baseline=-6pt]{\square[0.6][0.3]{k}{j}{s_1}}
   } {
                \tikz[baseline=-6pt]{\trapezoiddd[0.6][0.6][0.3]{i}{j}{
                    s_1+s_2+s^{\text{arc}}_{i,j}}{+ s^{\text{sib}}_{i, k, j}}}}
                    $
	      \end{tabular}
	      }
	    \end{subfigure}
    \caption{
    Deduction rules for our modified Eisner algorithm \cite{eisner-1997-bilexical} (first two rows) and its second-order extension \cite{mcdonald-pereira-2006-online} (all rows). Our modifications are highlighted in red. All deduction items are annotated with their scores.  Note that ``finished'' (in)complete spans are marked by double underlines. 
    }
    \label{fig:algo2}

\end{figure*}

\subsection{$O(n^4)$ modified Eisner-Satta algorithm}
\label{parsing1}
In this case, we combine first-order graph-based parsing and headed-span-based parsing. The score of a dependency tree $y$ is defined as:
\[
s(y) = \sum_{(x_i \rightarrow x_j) \in y} s^{\text{arc}}_{i,j} + \sum_{(l_i, r_i, x_i) \in y} s^{\text{span}}_{l_i, r_i, i}
\]

We adapt the Eisner-Satta algorithm for parsing. The $O(n^4)$ Eisner-Satta algorithm \cite[][{Sec. 3}]{eisner-satta-1999-efficient} is originally defined with bilexicalized PCFGs. Still, we can leverage its dynamic programming substructure to incorporate both arc scores and headed span scores, similar to the relationship between span-based constituency parsing \cite{stern-etal-2017-minimal} and PCFG parsing.    The axiom items are $\tikz[baseline=-10pt]{\triangle[0.6][0.6]{i}{i}{i}{\phantom{0}}}$ with initial score 0 and the deduction rules are listed in Fig. \ref{fig:algo}.  Unlike the original Eisner-Satta algorithm, we distinguish between ``finished'' spans and ``unfinished'' spans.  An ``unfinished'' span can absorb a child span to form a larger span, while in a ``finished'' span, the headword has already generated all its children, so it cannot expand anymore and corresponds to a headed-span for the given headword. By explicitly distinguishing between ``unfinished`` spans and ``finished`` spans, we can incorporate headed-span scores $s^{\text{span}}$ into parsing via the newly introduced rule \texttt{FINISH}.  We then modify the rule \texttt{L-LINK} and \texttt{R-LINK} accordingly as only a ``finished'' span can be attached. 

\subsection{$O(n^3)$ modifed Eisner algorithm}
\label{parsing2}
In order to decrease the parsing time complexity from $O(n^4)$ to $O(n^3)$, we decompose $s^{\text{span}}_{l,r,i}$ into two terms: 
\[
s(y) = \sum_{(x_i \rightarrow x_j) \in y} s^{\text{arc}}_{i,j} + \sum_{(l_i, r_i, x_i) \in y} (s^{\text{left}}_{i, l_i} + s^{\text{right}}_{i, r_i})
\]
and modify the Eisner algorithm accordingly. The axiom items are  $\tikz[baseline=-10pt]{\righttrianglee[0.6][0.4]{i}{i}{}}$ and $\tikz[baseline=-10pt]{\lefttrianglee[0.6][0.4]{i}{i}{}}$
  with initial score 0 and the deduction rules are shown in the first two rows of Fig. \ref{fig:algo2}.  Similar to the case in the previous subsection,  the original Eisner algorithm does not distinguish between ``finished'' complete spans and ``unfinished'' complete spans. An ``unfinished'' complete span can absorb another complete span to form a larger incomplete span, while a ``finished'' complete span has no more child in the given direction and thus cannot expand anymore. We introduce new rules \texttt{L-FINISH} and \texttt{R-FINISH} to incorporate the left or right span boundary scores respectively, and adjust other rules accordingly.

\begin{table*}[tb!]\small
	\centering
	\vskip -.0in
	{\setlength{\tabcolsep}{.8em}
		\makebox[\linewidth]{\resizebox{\linewidth}{!}{%
				\begin{tabular}{lcccccccccccc|l}
					\toprule
					  &	bg &	 ca & cs& de& en& es& fr & it&nl&no&ro&ru &Avg\\
					 \toprule
												\multicolumn{13}{c}{+$\text{BERT}_{\text{multilingual}}$}   \\
					\toprule
					$\textit{Biaffine+MM}^{\dagger}$ & 90.30 & 94.49 & 92.65 & \textbf{85.98} & 91.13 &93.78 & 91.77 &94.72 &91.04 &94.21 &87.24 & 94.53 &91.82 \\  
                     {\it Span} & 91.10 & 94.46 &92.57 &85.87 & \textbf{91.32} & 93.84 & 91.69 & 94.78 & 91.65 & 94.28 & 87.48 &94.45 & 91.96 \\
                     {\it 1O+Span} & 91.44 &94.54 &92.68 &85.75 &91.23 &93.84 &91.67 & \textbf{94.97} & \textbf{91.81} &94.35 & 87.17 &94.49 &91.99\\
                     {\it 1O+Span+Headsplit} &91.46 &94.53 &92.63 &85.78 &91.25 &93.77 & \textbf{91.91} &94.88 &91.59 & 94.18 &87.45 &94.47 &91.99\\ 
                    \midrule
                    {\it Biaffine+2O+MM} &  91.58 & 94.48  & \textbf{92.69} & 85.72& 91.28  & 93.80 & 91.89 & 94.88  & 91.30 & 94.23  & 87.55 & \textbf{94.55} & 92.00\\ 
                     {\it 2O+Span+Headsplit} & \textbf{91.82} & \textbf{94.58} &92.59 &85.65 &91.28 & \textbf{93.86} &91.80 &94.75 &91.50 & \textbf{94.40} & \textbf{87.71} &94.51 & \textbf{92.04}\\
                     \midrule 
                     \multicolumn{13}{c}{For reference}   \\
                     \midrule
                     				    {\it MFVI2O} & 91.30 & 93.60 & 92.09 & 82.00 & 90.75 & 92.62 & 89.32 & 93.66 & 91.21 & 91.74 & 86.40 & 92.61  & 90.61 \\ 

					\bottomrule				
	\end{tabular}}}}
	\caption{
		\label{ud_2.2} Labeled Attachment Score (LAS) on twelves languages in UD 2.2. We use ISO 639-1 codes to
represent languages.  $\dagger$ means reported by \citet{Yang2022Span}. MFVI2O: \citet{wang-tu-2020-second}. Span: \citet{Yang2022Span}. 
	}
	\label{result_ud} 
	\vskip -.12in
\end{table*}
\begin{table}[tb!]
    \centering 
    \small
    \begin{tabular}{lcccc}
        \toprule 
        & \multicolumn{2}{c}{{\bf PTB}} & \multicolumn{2}{c}{{\bf CTB}}\\
        & UAS & LAS & UAS & LAS \\
        \midrule 
        &\multicolumn{2}{c}{\underline{+$\text{BERT}_{\text{large}}$}}&\multicolumn{2}{c}{\underline{+$\text{BERT}_{\text{base}}$}}\\
              $\textit{Biaffine+MM}^{\dagger}$ & 97.22 & 95.71 & 93.18 & 92.10  \\
       {\it Span} & 97.24 & 95.73 & 93.33 & 92.30 \\ 
       {\it 1O+Span} & 97.26 & 95.68 & 93.56 & \textbf{92.49}  \\
       {\it 1O+Span+HeadSplit} & \textbf{97.30} &  \textbf{95.77} & 93.46& 92.42 \\ 
       \hline 
       	{\it Biaffine+2O+MM} & 97.28 & 95.73 & 93.42 & 92.34 \\ 
       {\it 2O+Span+HeadSplit} & 97.23 & 95.69 & \textbf{93.57} & 92.47\\
       \midrule
    &\multicolumn{4}{l}{For reference}\\
    \midrule
        {\it MFVI2O} & 96.91 & 95.34 & 92.55 & 91.69 \\
       {\it HierPtr} & 97.01 & 95.48 & 92.65 & 91.47 \\ 
               &\multicolumn{2}{c}{\underline{+$\text{XLNet}_{\text{large}}$}}&\multicolumn{2}{c}{\underline{+$\text{BERT}_{\text{base}}$}}\\
        {\it HPSG$^\flat$} & 97.20 & 95.72 & - & - \\
        {\it HPSG+LAL$^\flat$} & 97.42 & 96.26 & 94.56 & 89.28 \\
       \bottomrule 
    \end{tabular}

    \caption{Results on PTB and CTB. $\flat$ denotes use of additional constituency tree data and thus not comparable to our work. ${\dagger}$ denotes results reported by \citet{Yang2022Span}. HPSG: \citet{zhou-zhao-2019-head}; HPSG+LAL: \citet{mrini-etal-2020-rethinking}; HierPtr: \citet{DBLP:journals/corr/abs-2105-09611}.}
    \label{tab:ptb_ctb}
\end{table}

\subsection{$O(n^3)$ modified second-order Eisner algorithm}
\label{parsing3}
We further enhance the model with adjacent sibling information:
\begin{align*}
	s(y) = &  \sum_{(x_i \rightarrow x_j) \in y} s^{\text{arc}}_{i,j} + \sum_{(x_i \rightarrow \{x_j, x_k\})\in y} s^{\text{sib}}_{i,j,k}\\
      &	  + \sum_{(l_i, r_i, x_i) \in y} (s^{\text{left}}_{i, l_i} + s^{\text{right}}_{i, r_i})
\end{align*}
where for each adjacent sibling part $x_i \rightarrow \{x_j, x_k\}$, $x_j$ and $x_k$ are two adjacent dependents of $x_i$.

Similarly, we modify the second-order extension of the Eisner algorithm \cite{mcdonald-pereira-2006-online} by distinguishing between ``unfinished'' and ``finished'' complete spans.  The additional deductive rules for second-order parsing are shown in the last row of Fig. \ref{fig:algo2} and the length of the ``unfinished'' complete span is forced to be 1 in the rule \texttt{L-LINK} and \texttt{R-LINK}.

\section{Experiments}
\subsection{Setup}
We conduct experiments on in Penn Treebank (PTB) 3.0 \cite{marcus-etal-1993-building}, Chinese Treebank (CTB) 5.1 \cite{DBLP:journals/nle/XueXCP05} and 12 languages on Universal Dependencies (UD) 2.2. Implementation details are shown in \cref{implementation}. The reported results are averaged over three runs with different random seeds. 

\subsection{Main result}
Table \ref{result_ud} and \ref{tab:ptb_ctb} show the results on UD, PTB and CTB respectively. We additionally reimplement \texttt{Biaffine+2O+MM} by replacing the TreeCRF loss of \citet{zhang-etal-2020-efficient} with the max-margin loss for fair comparison. We refer to our proposed models as \texttt{1O+Span} (\cref{parsing1}), \texttt{1O+Span+Headsplit} (\cref{parsing2}), and \texttt{2O+Span+Headsplit} (\cref{parsing3}) respectively. 

We draw the following observations. (1) Second-order information is still helpful even with powerful encoders (i.e., BERT). \texttt{Biaffine+2O+MM} outperforms \texttt{Biaffine+MM} in almost all cases. (2)  Combining first-order graph-based and headed-span-based methods is effective. Both \texttt{1O+Span} and \texttt{1O+Span+Headsplit} beat \texttt{Biaffine+MM}, \texttt{Span} in almost all cases; have similar performance to \texttt{Biaffine+2O+MM}. (3) Decomposing the headed-span scores is useful. \texttt{1O+Span+Headsplit} has similar performance to \texttt{1O+Span} while manages to decrease the parsing complexity from $O(n^4)$ to $O(n^3)$. We speculate that powerful encoders mitigate the issue of independent scoring. (4) Combining second-order graph-based and headed-span-based methods has marginal effects.
 We speculate that the utility of adjacent sibling information and headed span information is overlapping.

\subsection{Error analysis}
Following \cite{mcdonald-nivre-2011-analyzing}, we plot UAS as a function of sentence length; F1 scores as functions of distance to root and dependency length on the CTB test set. We also follow \cite{Yang2022Span} to plot F1 score as a function of span length.

\begin{figure}[tb!]
\centering
	\begin{subfigure}[t]{0.48\linewidth}
		 \resizebox{\textwidth}{!}{%
\begin{tikzpicture}
\begin{axis}[
    legend style={
            at={(1,1)},anchor=north east
        },
    symbolic x coords={1-9,10-19,20-29,30-39,$\ge$40},
    xtick=data,
        xlabel={Sentence length},
        ylabel={UAS (100\%)}
    ]
        \addplot[ultra thick, color=red] coordinates {
(1-9, 93.07)
 (10-19, 95.16)
 (20-29, 93.84)
 (30-39, 93.52)
 ($\ge$40, 92.58)};

    \addlegendentry{Span}
    \addplot[ultra thick, color=blue] coordinates {
(1-9, 93.25)
 (10-19, 95.67)
 (20-29, 93.84)
 (30-39, 92.87)
 ($\ge$40, 92.31)};
    \addlegendentry{Biaffine+MM}
    \addplot[ultra thick, color=green] coordinates {
(1-9, 93.50)
 (10-19, 95.63)
 (20-29, 94.26)
 (30-39, 93.00)
 ($\ge$40, 92.93)};

    \addlegendentry{1O+Span}
\end{axis}
\end{tikzpicture}
}
\caption{}
\label{error_a}
\end{subfigure}
	\begin{subfigure}[t]{0.48\linewidth}
		 \resizebox{\textwidth}{!}{%
\begin{tikzpicture}

\begin{axis}[
    legend style={
            at={(1,.3)},anchor=north east
        },
    symbolic x coords={ROOT,1,2,3,4,5, 6, $\ge$7},
    xtick=data,
    xlabel={Distance to root},
    ylabel={F1 score (100\%)}
]

    \addplot[ultra thick, color=red] coordinates {
(ROOT, 90.31)
 (1, 91.89)
 (2, 93.62)
 (3, 94.17)
 (4, 94.07)
 (5, 93.95)
 (6, 93.64)
 ($\ge$7, 92.59)
 };

    \addlegendentry{Span}
    \addplot[ultra thick, color=blue] coordinates {
(ROOT, 89.75)
 (1, 91.22)
 (2, 93.67)
 (3, 94.25)
 (4, 93.88)
 (5, 93.78)
 (6, 93.50)
 ($\ge$7, 91.92)
 };


    \addlegendentry{Biaffine+MM}
    
\addplot[ultra thick, color=green] coordinates {
(ROOT, 90.20)
 (1, 91.71)
 (2, 93.94)
 (3, 94.70)
 (4, 94.19)
 (5, 94.03)
 (6, 93.90)
 ($\ge$7, 92.73)
 };
    \addlegendentry{1O+span}
\end{axis}
\end{tikzpicture}
}
\caption{}
\label{error_b}
\end{subfigure}
	\begin{subfigure}[t]{0.48\linewidth}
		 \resizebox{\textwidth}{!}{%
\begin{tikzpicture}

\begin{axis}[
    legend style={
            at={(1,1)},anchor=north east
        },
    symbolic x coords={1,2,3,4,5, 6, 7, $\ge$8},
    xtick=data,
    xlabel={Dependency length},
    ylabel={F1 score (100\%)}
]

    \addplot[ultra thick, color=red] coordinates {
 (1, 97.44)
 (2, 93.73)
 (3, 92.20)
 (4, 91.29)
 (5, 89.63)
 (6, 89.00)
 (7, 87.82)
 ($\ge$8, 82.31)
 };

    \addlegendentry{Span}

    \addplot[ultra thick, color=blue] coordinates {
 (1, 97.42)
 (2, 93.58)
 (3, 91.90)
 (4, 91.36)
 (5, 90.00)
 (6, 89.06)
 (7, 87.01)
 ($\ge$8, 81.37)
 };


    \addlegendentry{Biaffine+MM}

        \addplot[ultra thick, green] coordinates {
 (1, 97.46)
 (2, 93.96)
 (3, 92.25)
 (4, 91.33)
 (5, 90.40)
 (6, 89.62)
 (7, 88.65)
 ($\ge$8, 83.09)
 };

\addlegendentry{1O+span}

\end{axis}
\end{tikzpicture}
}
\caption{}
\label{error_c}
\end{subfigure}
\begin{subfigure}[t]{0.48\linewidth}
		 \resizebox{\textwidth}{!}{%
\begin{tikzpicture}

\begin{axis}[
    legend style={
            at={(1,1)},anchor=north east,
        },
    symbolic x coords={1-10,11-20,21-30,31-40,$\ge$40},
    xtick=data,
    xlabel={Span length},
    ylabel={F1 score (100\%)},
]

    \addplot[ultra thick, color=red] coordinates {
 (1-10, 96.31)
 (11-20, 86.3)
 (21-30, 83.74)
   (31-40, 83.59)
 ($\ge$40, 83.12)
 };
    \addlegendentry{Span}

    \addplot[ultra thick, color=blue] coordinates {
 (1-10, 96.14)
 (11-20, 85.39)
 (21-30, 83.28)
 (31-40, 81.56)
 ($\ge$40, 81.06)
 };

    \addlegendentry{Biaffine+MM}
    
    \addplot[ultra thick, color=green] coordinates {
 (1-10, 96.37)
 (11-20, 86.71)
 (21-30, 83.79)
 (31-40, 81.02)
 ($\ge$40, 82.11)
 };
     \addlegendentry{1O+Span}

\end{axis}
\end{tikzpicture}
}
\caption{}
\label{error_d}
\end{subfigure}

\caption{Error analysis on the CTB test set.}
\label{error_ctb}

\end{figure}
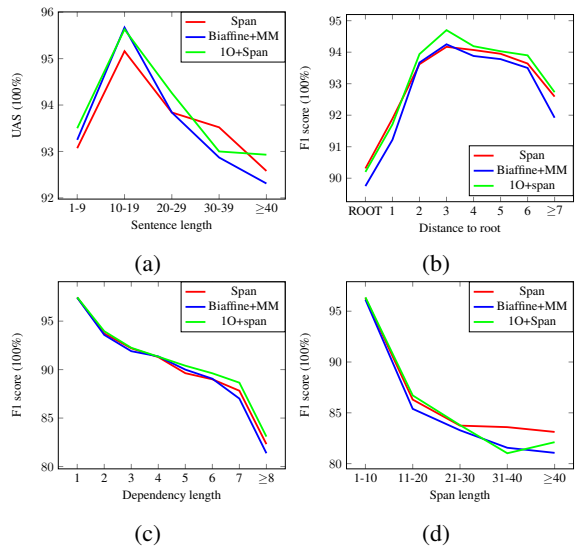

Fig. \ref{error_a} shows that compared with first-order graph-based method (i.e., Biaffine+MM),  headed-span-based method (i.e., Span) has an advantage in predicting long sentences (of length > 30) but has a difficulty in predicting short sentences (of length < 20). By combining first-order graph-based and headed-span-based methods, \texttt{1O+Span} can predict both short and long sentences correctly. It achieves the best results for all sentence length intervals except for 30-39.  Fig. \ref{error_b} shows that 1O+Span achieves the best performance for almost all cases, indicating its strong ability in predicting complex subtrees with high tree depth. Also, Fig. \ref{error_c} shows that 1O+Span achieves the best performance for almost all cases, especially for dependency arcs of length $\geq$ 6, showing its ability in capturing long-range dependencies. Fig. \ref{error_d} shows that \texttt{Span} has the best performance in identifying the range of a subtree, although it has no direct relation to the final performance.

\section{Conclusion}
In this paper, we have studied different ways to combine graph-based and headed-span-based methods. We found that applying the head-splitting trick can retain the cubic parsing complexity and meanwhile improve parsing performance when combining first-order graph-based and headed-span-based methods. We also confirmed the effectiveness of second-order parsing, however, we observed marginal or no improvement when combining it with the headed-span-based method.

\section*{Acknowledgments}
We thank the anonymous reviewers for their constructive comments. This work was supported by the National Natural Science Foundation of China (61976139).

\bibliography{anthology,custom}

\begin{thebibliography}{26}
\expandafter\ifx\csname natexlab\endcsname\relax\def\natexlab#1{#1}\fi

\bibitem[{Carreras(2007)}]{carreras-2007-experiments}
Xavier Carreras. 2007.
\newblock \href {https://aclanthology.org/D07-1101} {Experiments with a
  higher-order projective dependency parser}.
\newblock In \emph{Proceedings of the 2007 Joint Conference on Empirical
  Methods in Natural Language Processing and Computational Natural Language
  Learning ({EMNLP}-{C}o{NLL})}, pages 957--961, Prague, Czech Republic.
  Association for Computational Linguistics.

\bibitem[{Devlin et~al.(2019)Devlin, Chang, Lee, and
  Toutanova}]{devlin-etal-2019-bert}
Jacob Devlin, Ming-Wei Chang, Kenton Lee, and Kristina Toutanova. 2019.
\newblock \href {https://doi.org/10.18653/v1/N19-1423} {{BERT}: Pre-training of
  deep bidirectional transformers for language understanding}.
\newblock In \emph{Proceedings of the 2019 Conference of the North {A}merican
  Chapter of the Association for Computational Linguistics: Human Language
  Technologies, Volume 1 (Long and Short Papers)}, pages 4171--4186,
  Minneapolis, Minnesota. Association for Computational Linguistics.

\bibitem[{Dozat and Manning(2017)}]{DBLP:conf/iclr/DozatM17}
Timothy Dozat and Christopher~D. Manning. 2017.
\newblock \href {https://openreview.net/forum?id=Hk95PK9le} {Deep biaffine
  attention for neural dependency parsing}.
\newblock In \emph{5th International Conference on Learning Representations,
  {ICLR} 2017, Toulon, France, April 24-26, 2017, Conference Track
  Proceedings}. OpenReview.net.

\bibitem[{Eisner(1997)}]{eisner-1997-bilexical}
Jason Eisner. 1997.
\newblock \href {https://aclanthology.org/1997.iwpt-1.10} {Bilexical grammars
  and a cubic-time probabilistic parser}.
\newblock In \emph{Proceedings of the Fifth International Workshop on Parsing
  Technologies}, pages 54--65, Boston/Cambridge, Massachusetts, USA.
  Association for Computational Linguistics.

\bibitem[{Eisner and Satta(1999)}]{eisner-satta-1999-efficient}
Jason Eisner and Giorgio Satta. 1999.
\newblock \href {https://doi.org/10.3115/1034678.1034748} {Efficient parsing
  for bilexical context-free grammars and head automaton grammars}.
\newblock In \emph{Proceedings of the 37th Annual Meeting of the Association
  for Computational Linguistics}, pages 457--464, College Park, Maryland, USA.
  Association for Computational Linguistics.

\bibitem[{Falenska and Kuhn(2019)}]{falenska-kuhn-2019-non}
Agnieszka Falenska and Jonas Kuhn. 2019.
\newblock \href {https://doi.org/10.18653/v1/P19-1012} {The (non-)utility of
  structural features in {B}i{LSTM}-based dependency parsers}.
\newblock In \emph{Proceedings of the 57th Annual Meeting of the Association
  for Computational Linguistics}, pages 117--128, Florence, Italy. Association
  for Computational Linguistics.

\bibitem[{Fern{\'{a}}ndez{-}Gonz{\'{a}}lez and
  G{\'{o}}mez{-}Rodr{\'{\i}}guez(2021)}]{DBLP:journals/corr/abs-2105-09611}
Daniel Fern{\'{a}}ndez{-}Gonz{\'{a}}lez and Carlos
  G{\'{o}}mez{-}Rodr{\'{\i}}guez. 2021.
\newblock \href {http://arxiv.org/abs/2105.09611} {Dependency parsing with
  bottom-up hierarchical pointer networks}.
\newblock \emph{CoRR}, abs/2105.09611.

\bibitem[{Fonseca and Martins(2020)}]{fonseca-martins-2020-revisiting}
Erick Fonseca and Andr{\'e} F.~T. Martins. 2020.
\newblock \href {https://doi.org/10.18653/v1/2020.acl-main.776} {Revisiting
  higher-order dependency parsers}.
\newblock In \emph{Proceedings of the 58th Annual Meeting of the Association
  for Computational Linguistics}, pages 8795--8800, Online. Association for
  Computational Linguistics.

\bibitem[{Hochreiter and Schmidhuber(1997)}]{hochreiter1997long}
Sepp Hochreiter and J{\"u}rgen Schmidhuber. 1997.
\newblock Long short-term memory.
\newblock \emph{Neural computation}, 9(8):1735--1780.

\bibitem[{Kingma and Ba(2015)}]{DBLP:journals/corr/KingmaB14}
Diederik~P. Kingma and Jimmy Ba. 2015.
\newblock \href {http://arxiv.org/abs/1412.6980} {Adam: {A} method for
  stochastic optimization}.
\newblock In \emph{3rd International Conference on Learning Representations,
  {ICLR} 2015, San Diego, CA, USA, May 7-9, 2015, Conference Track
  Proceedings}.

\bibitem[{Koo and Collins(2010)}]{koo-collins-2010-efficient}
Terry Koo and Michael Collins. 2010.
\newblock \href {https://aclanthology.org/P10-1001} {Efficient third-order
  dependency parsers}.
\newblock In \emph{Proceedings of the 48th Annual Meeting of the Association
  for Computational Linguistics}, pages 1--11, Uppsala, Sweden. Association for
  Computational Linguistics.

\bibitem[{Ma and Zhao(2012)}]{ma-zhao-2012-fourth}
Xuezhe Ma and Hai Zhao. 2012.
\newblock \href {https://aclanthology.org/C12-2077} {Fourth-order dependency
  parsing}.
\newblock In \emph{Proceedings of {COLING} 2012: Posters}, pages 785--796,
  Mumbai, India. The COLING 2012 Organizing Committee.

\bibitem[{Marcus et~al.(1993)Marcus, Santorini, and
  Marcinkiewicz}]{marcus-etal-1993-building}
Mitchell~P. Marcus, Beatrice Santorini, and Mary~Ann Marcinkiewicz. 1993.
\newblock \href {https://aclanthology.org/J93-2004} {Building a large annotated
  corpus of {E}nglish: The {P}enn {T}reebank}.
\newblock \emph{Computational Linguistics}, 19(2):313--330.

\bibitem[{McDonald et~al.(2005)McDonald, Crammer, and
  Pereira}]{mcdonald-etal-2005-online}
Ryan McDonald, Koby Crammer, and Fernando Pereira. 2005.
\newblock \href {https://doi.org/10.3115/1219840.1219852} {Online large-margin
  training of dependency parsers}.
\newblock In \emph{Proceedings of the 43rd Annual Meeting of the Association
  for Computational Linguistics ({ACL}{'}05)}, pages 91--98, Ann Arbor,
  Michigan. Association for Computational Linguistics.

\bibitem[{McDonald and Nivre(2011)}]{mcdonald-nivre-2011-analyzing}
Ryan McDonald and Joakim Nivre. 2011.
\newblock \href {https://doi.org/10.1162/coli_a_00039} {Analyzing and
  integrating dependency parsers}.
\newblock \emph{Computational Linguistics}, 37(1):197--230.

\bibitem[{McDonald and Pereira(2006)}]{mcdonald-pereira-2006-online}
Ryan McDonald and Fernando Pereira. 2006.
\newblock \href {https://aclanthology.org/E06-1011} {Online learning of
  approximate dependency parsing algorithms}.
\newblock In \emph{11th Conference of the {E}uropean Chapter of the Association
  for Computational Linguistics}, Trento, Italy. Association for Computational
  Linguistics.

\bibitem[{Mrini et~al.(2020)Mrini, Dernoncourt, Tran, Bui, Chang, and
  Nakashole}]{mrini-etal-2020-rethinking}
Khalil Mrini, Franck Dernoncourt, Quan~Hung Tran, Trung Bui, Walter Chang, and
  Ndapa Nakashole. 2020.
\newblock \href {https://doi.org/10.18653/v1/2020.findings-emnlp.65}
  {Rethinking self-attention: Towards interpretability in neural parsing}.
\newblock In \emph{Findings of the Association for Computational Linguistics:
  EMNLP 2020}, pages 731--742, Online. Association for Computational
  Linguistics.

\bibitem[{Pereira and Warren(1983)}]{pereira-warren-1983-parsing}
Fernando C.~N. Pereira and David H.~D. Warren. 1983.
\newblock \href {https://doi.org/10.3115/981311.981338} {Parsing as deduction}.
\newblock In \emph{21st Annual Meeting of the Association for Computational
  Linguistics}, pages 137--144, Cambridge, Massachusetts, USA. Association for
  Computational Linguistics.

\bibitem[{Stern et~al.(2017)Stern, Andreas, and
  Klein}]{stern-etal-2017-minimal}
Mitchell Stern, Jacob Andreas, and Dan Klein. 2017.
\newblock \href {https://doi.org/10.18653/v1/P17-1076} {A minimal span-based
  neural constituency parser}.
\newblock In \emph{Proceedings of the 55th Annual Meeting of the Association
  for Computational Linguistics (Volume 1: Long Papers)}, pages 818--827,
  Vancouver, Canada. Association for Computational Linguistics.

\bibitem[{Taskar et~al.(2004)Taskar, Klein, Collins, Koller, and
  Manning}]{taskar-etal-2004-max}
Ben Taskar, Dan Klein, Mike Collins, Daphne Koller, and Christopher Manning.
  2004.
\newblock \href {https://aclanthology.org/W04-3201} {Max-margin parsing}.
\newblock In \emph{Proceedings of the 2004 Conference on Empirical Methods in
  Natural Language Processing}, pages 1--8, Barcelona, Spain. Association for
  Computational Linguistics.

\bibitem[{Taskar et~al.(2005)Taskar, Chatalbashev, Koller, and
  Guestrin}]{DBLP:conf/icml/TaskarCKG05}
Benjamin Taskar, Vassil Chatalbashev, Daphne Koller, and Carlos Guestrin. 2005.
\newblock \href {https://doi.org/10.1145/1102351.1102464} {Learning structured
  prediction models: a large margin approach}.
\newblock In \emph{Machine Learning, Proceedings of the Twenty-Second
  International Conference {(ICML} 2005), Bonn, Germany, August 7-11, 2005},
  volume 119 of \emph{{ACM} International Conference Proceeding Series}, pages
  896--903. {ACM}.

\bibitem[{Wang and Tu(2020)}]{wang-tu-2020-second}
Xinyu Wang and Kewei Tu. 2020.
\newblock \href {https://aclanthology.org/2020.aacl-main.12} {Second-order
  neural dependency parsing with message passing and end-to-end training}.
\newblock In \emph{Proceedings of the 1st Conference of the Asia-Pacific
  Chapter of the Association for Computational Linguistics and the 10th
  International Joint Conference on Natural Language Processing}, pages 93--99,
  Suzhou, China. Association for Computational Linguistics.

\bibitem[{Xue et~al.(2005)Xue, Xia, Chiou, and
  Palmer}]{DBLP:journals/nle/XueXCP05}
Naiwen Xue, Fei Xia, Fu{-}Dong Chiou, and Martha Palmer. 2005.
\newblock \href {https://doi.org/10.1017/S135132490400364X} {The penn chinese
  treebank: Phrase structure annotation of a large corpus}.
\newblock \emph{Nat. Lang. Eng.}, 11(2):207--238.

\bibitem[{Yang and Tu(2022)}]{Yang2022Span}
Songlin Yang and Kewei Tu. 2022.
\newblock Headed-span-based projective dependency parsing.
\newblock In \emph{ACL}.

\bibitem[{Zhang et~al.(2020)Zhang, Li, and Zhang}]{zhang-etal-2020-efficient}
Yu~Zhang, Zhenghua Li, and Min Zhang. 2020.
\newblock \href {https://doi.org/10.18653/v1/2020.acl-main.302} {Efficient
  second-order {T}ree{CRF} for neural dependency parsing}.
\newblock In \emph{Proceedings of the 58th Annual Meeting of the Association
  for Computational Linguistics}, pages 3295--3305, Online. Association for
  Computational Linguistics.

\bibitem[{Zhou and Zhao(2019)}]{zhou-zhao-2019-head}
Junru Zhou and Hai Zhao. 2019.
\newblock \href {https://doi.org/10.18653/v1/P19-1230} {{H}ead-{D}riven
  {P}hrase {S}tructure {G}rammar parsing on {P}enn {T}reebank}.
\newblock In \emph{Proceedings of the 57th Annual Meeting of the Association
  for Computational Linguistics}, pages 2396--2408, Florence, Italy.
  Association for Computational Linguistics.

\end{thebibliography}
\bibliographystyle{acl_natbib}

\appendix
\section{Implementation details}
\label{implementation}
 We use "bert-large-cased" for PTB, "bert-base-chinese" for CTB, and "bert-multilingual-cased" for UD.
   We set the  hidden size of BiLSTM  to 1000. We set the hidden size of  biaffine functions to 600/300 for spans,arcs/labels. We set the hidden size of triaffine functions to 400. We add a dropout layer after the embedding layer, LSTM layers, and MLP layers with dropout rate 0.33. We use Adam \cite{DBLP:journals/corr/KingmaB14} as the optimizer with $\beta_1 = 0.9, \beta_2=0.9$ to train our model for 10 epochs with gradient clipping of 5. 
   The maximal learning rate is $lr=5e-5$ for BERT and $lr=25e-5$ for other components. We linearly warmup the learning rate to their maximal value for the first epoch and gradually decay them to zero for the rest of the epochs.  We batch sentences of similar lengths so that the token number is 4000 for each batch.

\end{document}